%% file: MAIN.tex
\documentclass[letterpaper, 10pt, conference]{ieeeconf}  

\IEEEoverridecommandlockouts                              

\usepackage{amsmath, amssymb}

\usepackage{mathtools}
\usepackage{algorithm}
\usepackage{algorithmic}
\usepackage{bm}
\usepackage{cite}
\usepackage[dvipsnames]{xcolor}
\usepackage{optidef}
\usepackage{cleveref}
\usepackage{graphicx}
\usepackage{subcaption}
\usepackage{svg}

\usepackage{mathtools}

\usepackage{etoolbox}

\newtheorem{remark}{Remark}

\newtheorem{definition}{Definition}

\title{\LARGE \bf
Multi-Agent Motion Planning on Industrial Magnetic Levitation Platforms: A Hybrid ADMM-HOCBF approach
}

\author{Bavo Tistaert$^{*}$, Stan Servaes$^{*}$, Alejandro Gonzalez-Garcia$^{**}$, Ibrahim Ibrahim$^{**}$, \\ Louis Callens, Jan Swevers and Wilm Decré 
\thanks{Authors are with MECO~Research~Team,~Dept.~of~Mechanical~Engineering,~KU~Leuven and Flanders Make @KU Leuven, 3001 Leuven, Belgium.
        {\tt\small firstname.name@kuleuven.be}}%
\thanks{This work is based on the Master's Thesis \cite{alma9995771124601488} and has been carried out within the framework of the Flanders Make SBO project ARENA: Agile and Reliable Navigation. Flanders Make is the Flemish strategic research centre for the manufacturing industry.}%
\thanks{$^{*}$B. Tistaert and S. Servaes are equal contributors to this work.}%
\thanks{$^{**}$A. Gonzalez-Garcia and I. Ibrahim are equal contributors to this work.}%
}

\begin{document}

\maketitle
\thispagestyle{empty}
\pagestyle{empty}

\begin{abstract}

This paper presents a novel hybrid motion planning method for holonomic multi-agent systems. The proposed decentralised model predictive control (MPC) framework tackles the intractability of classical centralised MPC for a growing number of agents while providing safety guarantees. This is achieved by combining a decentralised version of the alternating direction method of multipliers (ADMM) with a centralised high-order control barrier function (HOCBF) architecture. Simulation results show significant improvement in scalability over classical centralised MPC. We validate the efficacy and real-time capability of the proposed method by developing a highly efficient C++ implementation and deploying the resulting trajectories on a real industrial magnetic levitation platform.

\end{abstract}


\input{content/section1}
\input{content/section2}
\input{content/section3}
\input{content/section4}
\input{content/section5}
\input{content/section6}






{\small
\bibliographystyle{IEEEtran}
\bibliography{references}
}

\end{document}

%% file: content/section1.tex
\section{Introduction} \label{sec:introduction}

In recent years, trends in industry show a growing interest in multi-agent systems, where multiple agents, such as warehouse robots, autonomous ground vehicles, autonomous surface vessels, or unmanned aerial vehicle swarms~\cite{li2021lifelong, ruben2017decentralised, multipleboats, luis2019ondemand}, perform complex tasks that are beyond the capabilities of a single agent, either by cooperating or by distributing multiple tasks over the agents. 

\begin{figure}
    \centering
    \includegraphics[width=0.9\linewidth]{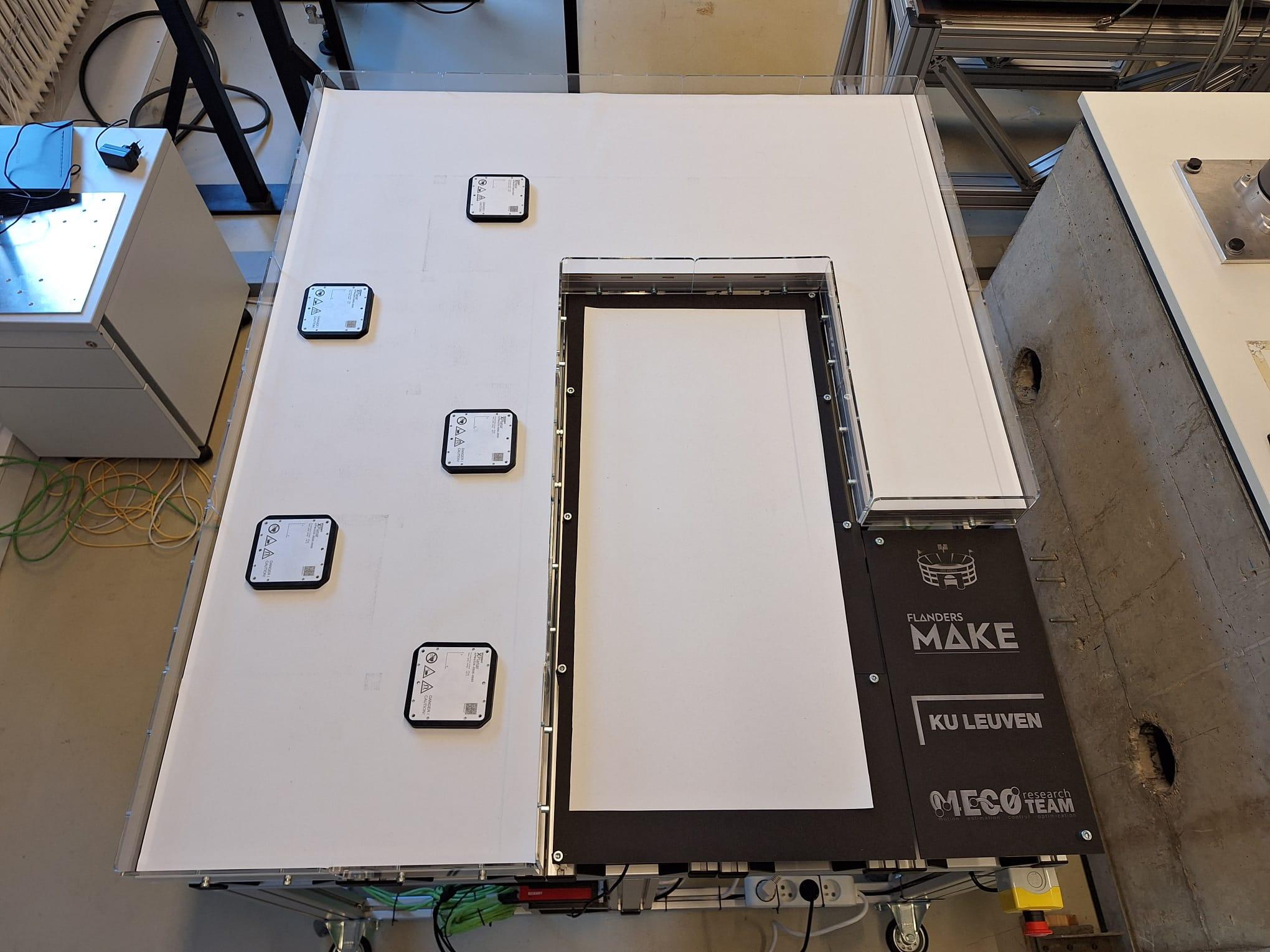}
    \caption{The Beckhoff XPlanar setup in the robotics lab. This industrial machine is used for the experimental validation of the motion planner.}
    \label{fig: xplanar experimental setup}
\end{figure}

An example of a system primed for multi-agent applications is the Beckhoff XPlanar~\cite{beckhoffwebsite} (Fig.~\ref{fig: xplanar experimental setup}), originally developed as an efficient, accurate, and precise transport system for production lines. Its working principle is based on magnetic levitation, where passive square-shaped movers, composed of permanent magnets, are actuated using active tiles, driving the movers anywhere on top of these tiles with very high precision and accuracy. The holonomic characteristic of the movers introduces great flexibility in their motion, but also provides many feasible paths for each mover. Hence, challenges persist in the coordination of agents in these systems. When restricting their motion to pre-defined tracks with traffic rules, coordination becomes much easier. However, this undermines the flexibility of the system and makes the production line susceptible to traffic jams, causing production delays. 

An appealing way to make use of this flexibility and thereby increase efficiency is to employ a real-time motion planner. Existing heuristic-based classical planners such as A*~\cite{hart1968Astar}, Rapidly exploring random trees (RRT)~\cite{lavalle1998RRT}, and artificial potential fields~\cite{khatib1986APF} and their variations are not ideally suited for the problem at hand. 
A*-class methods rely on discrete graphs, require graph-building, and often come with no safety or optimality guarantees. On the other hand, RRT variants are not guaranteed to find solutions and are more catered towards higher dimensional planning problems~\cite{zhang2025samplingbased}. Lastly, artificial potential fields notoriously suffer from local minima and deadlocks~\cite{koren1991potentialfieldmethods}. Most importantly, those methods are also catered to single-agent scenarios and must rely on additional planning schemes to properly perform multi-agent planning. In contrast, learning techniques are becoming increasingly popular and have been applied to large-scale multi-agent motion planning problems due to their favourable scalability. However, due to the lack of sufficient formal proofs in this area, they often have to be coupled with existing motion planning techniques \cite{qin2021learningcbf, zinage2024neuralcbfmpc}.

Alternatively, Model Predictive Control (MPC) naturally lends itself to multi-agent motion planning. MPC allows for flexible real-time motion planning under constraints in dynamic environments. Due to its inherent optimality, the resulting trajectories are very efficient, leading to reduced transit times.
The most efficient trajectories would be obtained with centralised MPC, which solves one central optimisation problem for all the agents. However, the computational cost for such problems often scales unfavourably with the number of agents, rendering the problem intractable for real-time motion planning. For multi-agent systems, this issue is often addressed by reformulating or approximating collision avoidance constraints~\cite{rey2018fullydecentralised, zhou2021cfsmpc, liu2018cfs, ruben2017decentralised, zhang2020exactconvexcollision}, by introducing neighbourhoods~\cite{rey2018fullydecentralised, chahine2023local, luis2019ondemand}, and by using distributed optimisation methods~\cite{shorinwa2024distributedtutorial}. This work focusses on the latter, which include distributed first-order methods, distributed sequential convex programming, and the alternating direction method of multipliers (ADMM). Here, the optimisation problem is decoupled and the sub-problems are distributed over the agents, with the possibility for parallel computations. We use a decentralised variation of ADMM due to its ability to fully decouple heavily constrained (convex) problems~\cite{ruben2017decentralised, ruben2016online, rey2018fullydecentralised}. 

Another important aspect in multi-agent motion planning is safety, \textit{i.e.}, agents must not collide with each other or with obstacles in the environment. When the original motion planner is not inherently safe, control barrier functions (CBFs) \cite{ames2019cbftheory} can provide a safety guarantee for systems with a relative degree of one. These techniques have therefore been applied to collision avoidance constraints in multi-agent systems~\cite{borrmann2015cbfsafeswarm, wang2017cbfsafety}, where they can also be decentralised. More recently, this concept was generalised to high-order control barrier functions (HOCBF)~\cite{xiao2021hocbf}, which apply to systems of any relative degree. Since ADMM is only guaranteed to converge for convex problems~\cite{boyd2011admm}, it cannot guarantee safety in nonconvex settings. Therefore, \cite{jiang2023incorporating} incorporates first-order discrete-time CBFs into the ADMM scheme to ensure safety. However, this strategy is only valid for systems of relative degree one.

Our work presents a hybrid multi-agent motion planner that complements decentralised ADMM by adding a separate centralised safety filter to the algorithm that uses continuous-time, second-order HOCBFs. Hence, this motion planner is safe for systems with acceleration input. Moreover, we leverage the centralised control architecture of the XPlanar by computing the safety filter in a centralised manner, which is known to result in more efficient and less conservative trajectories than computing it in a decentralised manner~\cite{borrmann2015cbfsafeswarm}.

In summary, the main contributions of this work are: 
\begin{itemize}
    \item A novel hybrid multi-agent motion planner which combines decentralised ADMM with centralised HOCBFs;
    \item A comparison of our proposed hybrid method against classical centralised MPC;
    \item Hardware experiments on a Beckhoff XPlanar industrial setup where we deploy computed trajectories.
\end{itemize}

The remainder of this paper is structured as follows: Section~\ref{sec:preliminaries} introduces the conventions, preliminary notions, and the problem formulation of this work. Section~\ref{sec:algorithm} then discusses the method and algorithm used to tackle this problem. The results are presented in Section~\ref{sec:experiments}, which contains extensive simulations to compare the novel method with the classic centralised MPC approach, as well as experiments on an industrial setup for a proof of concept. Finally, Section~\ref{sec:conclusion} draws some conclusions and suggests future work.

%% file: content/section2.tex
\section{Preliminaries and problem formulation} 
\label{sec:preliminaries}

After some conventions and definitions, this section describes and formulates the problem to be solved by the algorithm.

\subsection{Conventions}

Consider the problem of generating 2D trajectories for agents $i \in \{1,...,N\}$, each consisting of a set of positions $\textbf{p}_{i}$(t), velocities $\textbf{v}_{i}(t)$, and accelerations $\textbf{a}_{i}(t)$, where $t \in [0,T_{f}]$. Let an agent be modelled as as two decoupled double integrators with

\begin{equation}
\label{eq: Definition state vector}
\textbf{x}_{i}(t) = [p_{x,i}(t), \; p_{y,i}(t), \; v_{x,i}(t), \; v_{y,i}(t)]^{\top}    
\end{equation}
\begin{equation}
\label{eq: Definition input vector}
\textbf{u}_{i}(t) = [a_{x,i}(t), \; a_{y,i}(t)]^{\top},    
\end{equation}

\noindent
yielding:
\begin{equation}
    \dot{\textbf{x}}_{i}(t) = \textbf{A}\textbf{x}_{i}(t) + \textbf{B}\textbf{u}_{i}(t),
\end{equation}

\noindent
with
\begin{equation}
\label{eq: dynamics A and B}
    \textbf{A}
    =
    \begin{bmatrix}
        \textbf{0}_{2x2} & \textbf{I}_{2x2} \\
        \textbf{0}_{2x2} & \textbf{0}_{2x2}
    \end{bmatrix}
    \quad \textrm{and} \quad
    \textbf{B}
    =
    \begin{bmatrix}
        \textbf{0}_{2x2}\\
        \textbf{I}_{2x2}
    \end{bmatrix}.
\end{equation}

\subsection{High-Order Control Barrier Functions}
\label{sec:HOCBF}

\begin{definition}
A system is nonlinear control-affine if the state space can be described in the following form \cite{xiao2021hocbf}:
\begin{equation}
\label{eq: CBF nonlinear affine control system}
\dot{\textbf{x}} = \textbf{f}(\textbf{x}) + \textbf{g}(\textbf{x})\textbf{u} .
\end{equation}
\end{definition}

\noindent
The theory of HOCBFs considers systems of this form. Since system (\ref{eq: dynamics A and B}) is linear control-affine, it is a special case of system (\ref{eq: CBF nonlinear affine control system}), and thus the theory applies.

\begin{definition}
    ``A continuous function $\alpha: [0,a) \to [0, \infty), \: a > 0$ belongs to class $\mathcal{K}$ if it is strictly increasing and $\alpha(0) = 0$.'' \cite{xiao2021hocbf}
\end{definition}

\begin{definition}
    ``A set $\mathcal{C}$ is forward invariant for a system if its solutions starting at any $\textbf{x}(t_{0}) \in \mathcal{C}$ satisfy $\textbf{x}(t) \in \mathcal{C}$ for $\forall t > t_{0}$.'' \cite{xiao2021hocbf} 
\end{definition}

 \begin{definition}
     ``The relative degree $r$ of a (sufficiently) differentiable function $h$ with respect to system (\ref{eq: CBF nonlinear affine control system}) is the number of times we need to differentiate it along the dynamics of (\ref{eq: CBF nonlinear affine control system}) until the control $\textbf{u}$ explicitly shows.'' \cite{xiao2021hocbf}
 \end{definition}

When $h(\textbf{x}) \ge 0$ represents a safety constraint, rendering the superlevel set of $h$ invariant implies that the system will never leave this safe set, thereby imposing safety \cite{ames2019cbftheory}. When the relative degree of the system is larger than one, the following functions must be constructed \cite{xiao2021hocbf}:
\begin{equation}
\label{eq: HOCBF intermediate functions}
\begin{aligned}
    & \psi_{i}(\textbf{x}) = \dot \psi_{i-1}(\textbf{x}) + \alpha_{i}(\psi_{i-1}(\textbf{x})) \quad \forall i \in \{1,...,r\} \\
    & \text{with} \quad \psi_{0}(\textbf{x}) = h(\textbf{x}),
\end{aligned}
\end{equation}
where $\alpha_{i}$ denote class-$\mathcal{K}$ functions. Along with these functions, the following sets are defined:
\begin{equation}
\label{eq: HOCBF intermediate sets}
    \mathcal{C}_{i} = \{ \textbf{x} \in \mathbb{R}^{n_{x}} : \psi_{i-1}(\textbf{x}) \ge 0 \} \quad \forall i \in \{1,...,r\}.
\end{equation}
Due to the relative degree of $r$, only $\psi_{r}(\textbf{x}, \textbf{u})$ depends on the inputs. It can then be proven that the condition for rendering the safe set forward invariant is the following \cite{xiao2021hocbf}:
\begin{equation}
\label{eq: HOCBF invariance condition}
    \sup_{\textbf{u} \in U} \big [\psi_{r}(\textbf{x}, \textbf{u)} \big ] \ge 0 \quad \forall \textbf{x} \in \mathcal{C}_{1} \cap ...\cap \mathcal{C}_{r},
\end{equation}
which can be expanded to obtain the formal condition on $h$ to classify as a HOCBF \cite{xiao2021hocbf}:
\begin{align}
\label{eq: HOCBF general definition}
     & \sup_{\textbf{u} \in U} \bigg[ L_{\textbf{f}}^{r}h(\textbf{x}) + L_{\textbf{g}} L_{\textbf{f}}^{r-1}h(\textbf{x})\textbf{u} \notag + \sum_{i=1}^{r-1} L_{\textbf{f}}^{i} (\alpha_{r-i}(\psi_{r-i-1}(\textbf{x}))) \\ 
     & + \alpha_{r}(\psi_{r-1}(\textbf{x})) \bigg] \ge 0 \hspace{1cm} \forall \textbf{x} \in \mathcal{C}_{1} \cap \ldots \cap \mathcal{C}_{r}.
\end{align}
\noindent
Here, $L_{\textbf{f}}$ and $L_{\textbf{g}}$ denote the Lie derivatives along $\textbf{f}(\textbf{x})$ and $\textbf{g}(\textbf{x})$ respectively. Note that the expression within the brackets is linearly dependant on the input \textbf{u}, making it computationally efficient to use in optimisation problems.

\subsection{Problem formulation}

The goal of the motion planner is to steer every agent from an initial state to a target state, where the latter can be adapted in real-time for maximum flexibility. In order to generate optimal, collisionless trajectories, we rely on an MPC-based motion planner which solves the following optimal control problem (OCP) during every time step $\Delta t$:
\begin{subequations}
\label{eq: Centralised ocp}
\begin{align}
\min_{\forall \textbf{x}_{i}(\cdot), \textbf{u}_{i}(\cdot)} \quad & \int_{0}^{T_{f}} \sum_{i = 1}^{N} J_{i}(\textbf{x}_{i}(t), \textbf{u}_{i}(t)) \: dt \label{eq: centralised OCP obj}\\
\textrm{s.t.} \quad & \textbf{x}_{i}(0) = \textbf{x}_{0,i} \label{eq: centralised ocp init} \\
& \dot{\textbf{x}}_{i}(t) = \textbf{A}\textbf{x}_{i}(t) + \textbf{B}\textbf{u}_{i}(t) \\
& \lvert \lvert \textbf{v}_{i}(t) \rvert \rvert^{2} \le v^2_{max} \label{eq: centralised ocp max speed} \\
& \lvert \lvert \textbf{a}_{i}(t) \rvert \rvert^{2} \le a^2_{max} \label{eq: centralised ocp max acceleration} \\
& \lvert \lvert \textbf{p}_{i}(t) - \textbf{p}_{j}(t) \rvert \rvert^{2} \ge (2R + \epsilon)^2 \quad \forall j \ne i \label{eq: centralised MPC collision avoidacnce constraints} \\
& \forall \; i \in \{1,...,N\}, t \in [0,T_{f}]. \nonumber
\end{align}
\end{subequations}

\noindent
This formulation constitutes the classical centralised MPC approach. The objective function (\ref{eq: centralised OCP obj}) is the sum of each agent's own objective, which represents a quadratic penalisation from its target state $\textbf{x}_{f,i}$, along with an input penalisation for smoothness: 
\begin{subequations}
\label{eq: obj function}
\begin{align}
    J_{i}(\textbf{x}_{i}(t), \textbf{u}_{i}(t)) & = (\textbf{x}_{i}(t)-\textbf{x}_{f,i}(t))^{\top} \textbf{Q}_{i} (\textbf{x}_{i}(t)-\textbf{x}_{f,i}(t)) \nonumber\\
    & + \textbf{u}_{i}(t)^{\top} \textbf{R}_{i} \textbf{u}_{i}(t). 
\end{align}
\end{subequations}

\noindent
Eq.~(\ref{eq: centralised ocp init})-(\ref{eq: centralised ocp max acceleration}) respectively impose on each agent its initial condition, system dynamics, and an upper limit on the magnitude of its velocity and acceleration. Finally, Eq.~(\ref{eq: centralised MPC collision avoidacnce constraints}) represent the nonconvex collision avoidance constraints, which are enforced by imposing a minimal Euclidean distance between every pair of agents. Hence, agents are modelled as circles of radius $R$ in this constraint, and a safety margin $\epsilon$ is taken. Note that this circular model introduces conservatism since the physical agents are square-shaped. All constraints, with the exception of the initial condition, are enforced over the full time interval $[0,T_{f}]$. 

%% file: content/section3.tex
\section{Hybrid method and algorithm} \label{sec:algorithm}

This section discusses the method for solving OCP~(\ref{eq: Centralised ocp}) and how it is incorporated into the motion planning algorithm. First, we use decentralised ADMM to decouple and solve OCP~(\ref{eq: Centralised ocp}). As this method is not inherently safe, we complement it with an extra step which employs HOCBFs to recover the safety guarantee. Then, the pseudo-code for the motion planning algorithm is discussed. Finally, additional constraints are discussed to restrict the agents to a specified arena, e.g. the boundaries of the transport system.

\subsection{ADMM}

OCP~(\ref{eq: Centralised ocp}) is a coupled problem for all agents due to the collision avoidance constraints. Following \cite{ruben2017decentralised}, we may decouple the problem by reformulating constraints~(\ref{eq: centralised MPC collision avoidacnce constraints}) as follows:
\begin{subequations}
\label{eq: ADMM OCP}
\begin{align}
& \textbf{p}_{i}(t) = \textbf{z}_{i}(t), \; \textbf{p}_{j}(t) = \textbf{z}_{i,j}(t) \label{eq:ADMM slack vars} \quad j \in \mathcal{N}_{i} \\
& \lvert \lvert \textbf{z}_{i}(t) - \textbf{z}_{i,j}(t) \rvert \rvert^{2} \ge (2R + \epsilon)^2 \quad j \in \mathcal{N}_{i} \label{eq:ADMM coll constr}
\end{align}
\end{subequations}

\noindent
Here, we introduce for each agent $i$ the slack variables $\textbf{z}_{i}(t)$ and $\textbf{z}_{i,j}(t)$ along with the consensus constraints~(\ref{eq:ADMM slack vars}). The latter constraints ensure that the slack variables represent local copies of its own positions and the other agents' positions. This allows us to write the collision avoidance constraints~(\ref{eq:ADMM coll constr}) as a function of the local variables $\textbf{z}_{i}(t)$ and $\textbf{z}_{i,j}(t)$ of agent $i$ alone, instead of the positions $\textbf{p}_{i}(t)$ and $\textbf{p}_{j}(t)$ of agent $i$ and $j$. Moreover, neighbourhoods $\mathcal{N}_i$ are commonly introduced for each agent $i$ to decrease the computational load. Consequently, an agent only needs to enforce collision avoidance between other agents that pose a safety threat, e.g. only when they are within a certain detection radius from each other \cite{rey2018fullydecentralised}.

\begin{remark}\label{rem: all agents neighbours}
    Due to the highly dynamic nature of our problem, agent neighbourhoods become time-dependent, which standard ADMM cannot efficiently handle~\cite{shorinwa2024distributedtutorial}. Therefore, in this paper, each agent's neighbourhood is taken to contain all other agents.
\end{remark}

The augmented Lagrangian of the reformulated problem is obtained by only dualising the consensus constraints \cite{ruben2017decentralised}:
\begin{subequations}
\label{eq: Lagrangian ADMM}
\begin{align}
    \mathcal{L}_{\mu} & = \int_{0}^{T_f} \sum_{i=1}^{N} J_i(\mathbf{x}_i, \mathbf{u}_i) + \boldsymbol{\lambda}_i^{\top}(\mathbf{p}_i - \mathbf{z}_i) + \frac{\mu}{2} \|\mathbf{p}_i - \mathbf{z}_i\|^2 dt \nonumber \\
    & + \int_{0}^{T_f} \sum_{i=1}^{N} \sum_{j \in \mathcal{N}_i} \boldsymbol{\lambda}_{i,j}^{\top}(\mathbf{p}_j - \mathbf{z}_{i,j}) + \frac{\mu}{2} \|\mathbf{p}_j - \mathbf{z}_{i,j}\|^2 dt \label{eq: ADMM lagrangian} \\
    & = \sum_{i = 1}^{N} \mathcal{L}_{\mu,i}(\textbf{x}_{i}, \textbf{u}_{i}, \textbf{p}_{i}, \textbf{z}_{i}, \bm \lambda_{i}, \textbf{p}_{j}, \textbf{z}_{i,j}, \bm \lambda_{i,j}) \label{eq: ADMM sublagrangian} \\
    & = \sum_{i = 1}^{N} \mathcal{L}_{\mu,i}(\textbf{x}_{i}, \textbf{u}_{i}, \textbf{p}_{i}, \textbf{z}_{i}, \bm \lambda_{i}, \textbf{p}_{i}, \textbf{z}_{j, i}, \bm \lambda_{j, i}). \label{eq: ADMM sublagrangian reshuffled}
\end{align}
\end{subequations}

\noindent
Here, we can drop the time argument for brevity. Hence, the global Lagrangian $\mathcal{L}_{\mu}$ of OCP~(\ref{eq: Centralised ocp}) with constraints~(\ref{eq: ADMM OCP}) is the sum of every agent's local Lagrangian $\mathcal{L}_{\mu,i}$, for which two expressions can be formulated, Eq.~(\ref{eq: ADMM sublagrangian}) and (\ref{eq: ADMM sublagrangian reshuffled}). The equivalence between Eq.~(\ref{eq: ADMM sublagrangian}) and (\ref{eq: ADMM sublagrangian reshuffled}) is obtained by reshuffling the terms under the summation signs in Eq.~(\ref{eq: ADMM lagrangian}). This is only possible when the bidirectional identity holds \cite{ruben2017decentralised}, \textit{i.e.}, agent $j$ considers agent $i$ as neighbour if and only if agent $i$ considers agent $j$ as neighbour. By taking each agent's neighbourhood to contain all other agents (see Remark~\ref{rem: all agents neighbours}), this condition is automatically satisfied.

We then use decentralised ADMM to solve OCP~(\ref{eq: Centralised ocp}) with constraints~(\ref{eq: ADMM OCP}) in an iterative manner, first minimising Eq.~(\ref{eq: ADMM sublagrangian reshuffled}) w.r.t. $\textbf{x}_{i}$ and $\textbf{u}_{i}$, then minimising Eq.~(\ref{eq: ADMM sublagrangian}) w.r.t. $\textbf{z}_{i}$, $\textbf{z}_{i,j}$, and finally updating $\boldsymbol{\lambda}_{i}$ and $\boldsymbol{\lambda}_{i,j}$. Eq.~(\ref{eq: ruben ADMM general form OCPX, filled in})-(\ref{eq: ruben ADMM general form lambda update, filled in}) describe this sequence of computations. First, in OCP~(\ref{eq: ruben ADMM general form OCPX, filled in}), we only enforce each agent's own dynamic constraints and actuation limits, without considering collision avoidance. After this step is solved, each agent communicates its updated position $\textbf{p}_i^{k+1}$ with all its neighbours. Solving NLP~(\ref{eq: ruben ADMM general form OCPZ, filled in}) for each agent corrects these paths using the collision avoidance constraints. The results are stored as agent $i$'s local copies $\textbf{z}_i^{k+1}$ and $\textbf{z}_{i,j}^{k+1}$. These are subsequently communicated and serve as ``suggestions'' which can be used by the neighbours as a safe reference path. Lastly, the dual variables are updated using Eq.~(\ref{eq: ruben ADMM general form lambda update, filled in}). In the next iteration, OCP~(\ref{eq: ruben ADMM general form OCPX, filled in}) takes into account the local copies (or ``suggestions'') of the previous iteration through penalisation terms in the augmented Lagrangian.
\begin{equation}
\label{eq: ruben ADMM general form OCPX, filled in}
\begin{aligned}
\begin{pmatrix}
\textbf{x}_{i}^{k+1} \\
\textbf{u}_{i}^{k+1}
\end{pmatrix}\
= 
\:
\underset{\textbf{x}_{i}, \textbf{u}_{i}}{\arg \min} \quad & \mathcal{L}_{\mu,i}(\textbf{x}_{i}, \textbf{u}_{i}, \textbf{z}_{i}^{k}, \boldsymbol{\lambda}_{i}^{k}, \textbf{p}_{i}, \textbf{z}_{j,i}^{k}, \boldsymbol{\lambda}_{j,i}^{k}) \\
\textrm{s.t.} \quad  & \textbf{x}_{i}(0) = \textbf{x}_{0,i} \\
& \dot{\textbf{x}}_{i} = \textbf{A}\textbf{x}_{i} + \textbf{B}\textbf{u}_{i} \\
& \lvert \lvert \textbf{v}_{i} \rvert \rvert^{2} \le v^2_{max} \\
& \lvert \lvert \textbf{a}_{i} \rvert \rvert^{2} \le a^2_{max}  \\
\end{aligned}
\end{equation}
\newline
\begin{equation}
\label{eq: ruben ADMM general form OCPZ, filled in}
\begin{pmatrix}
\textbf{z}_{i}^{k+1}\\
\textbf{z}_{i,j}^{k+1}
\end{pmatrix}
=
\:
\begin{aligned}
& \underset{\textbf{z}_{i}, \textbf{z}_{i,j}}{\arg \min} \quad \mathcal{L}_{\mu,i}(\textbf{p}_{i}^{k+1}, \textbf{z}_{i}, \boldsymbol{\lambda}_{i}^{k}, \textbf{p}_{j}^{k+1}, \textbf{z}_{i,j}, \boldsymbol{\lambda}_{i,j}^{k}) \\
& \textrm{s.t.} \quad  \lvert \lvert \textbf{z}_{i} - \textbf{z}_{i,j} \rvert \rvert^{2} \ge (2R + \epsilon)^2 \quad j \in \mathcal{N}_{i}
\end{aligned}
\end{equation}
\newline
\begin{equation}
\label{eq: ruben ADMM general form lambda update, filled in}
\begin{aligned}
& \boldsymbol{\lambda}_{i}^{k+1} = \boldsymbol{\lambda}_{i}^{k} + \mu ({\textbf{p}}_{i}^{k+1} - \textbf{z}_{i}^{k+1}) \\
& \boldsymbol{\lambda}_{i,j}^{k+1} = \boldsymbol{\lambda}_{i,j}^{k} + \mu (\textbf{p}_{j}^{k+1} - \textbf{z}_{i,j}^{k+1}) \quad \forall j \in \mathcal{N}_{i}
\end{aligned}
\end{equation}

\noindent
This iterative procedure continues until the agents reach a consensus, \textit{i.e.}, we obtain optimal collisionless trajectories. In an MPC setting, the number of iterations per time step is often limited to avoid redundant computations~\cite{ruben2017decentralised}.

\begin{remark}
    Note that in updates~(\ref{eq: ruben ADMM general form OCPX, filled in})-(\ref{eq: ruben ADMM general form lambda update, filled in}), each agent's optimisation problem is decoupled. Computations for each agent can be carried out in parallel, with communication steps in between. 
\end{remark}

Since ADMM's convergence has only been proven for convex optimisation problems \cite{boyd2011admm}, there is no formal guarantee that the agents will reach a consensus due to the nonconvex collision avoidance constraints. Because safety is of crucial importance for industrial production lines, an additional step in the procedure is needed to recover a safety guarantee.

\subsection{ADMM-HOCBF}

To address the safety issue, we design an HOCBF to be used in an additional computationally-efficient step that enforces safety. We choose the left hand side of the collision avoidance constraint as HOCBF, after rewriting it in the following standard form:
\begin{equation}
\label{eq: collision avoidance constraint as cbf}
    h_{i,j} = \lvert \lvert \textbf{p}_{i}(0) - \textbf{p}_{j}(0) \rvert \rvert^{2} - (2R + \epsilon)^2 \geq 0.
\end{equation}

\noindent
Note that the HOCBF only acts at the current time step $t=0$. Since this function depends only on the positions of agents $i$ and $j$, the relative degree of $h_{i,j}$ is $r=2$. Therefore, we construct the following two functions:
\begin{subequations}
\label{eq: HOCBF collision avoidance intermediate functions}
\begin{align}
    & \psi_{1} = \dot h_{i,j} + \alpha_{1}(h_{i,j}) \label{eq: HOCBF collision avoidance intermediate functions psi 1}\\
    & \psi_{2} = \dot \psi_{1} + \alpha_{2}(\psi_{1}). \label{eq: HOCBF collision avoidance intermediate functions psi 2}
\end{align}

\end{subequations}

\noindent
We choose linear class-$\mathcal{K}$ functions \cite{xiao2021hocbf} for $\alpha_i(h) = K_i h$ to simplify the expressions. This results in the following condition that $h_{i,j}$ must satisfy in order to be a HOCBF:
\begin{equation}
\label{eq: condition on hij}
    \mathcal{H}_{i,j}(\textbf{u}_{i}(0), \textbf{u}_{j}(0)) = \ddot{h}_{i,j} + (K_{1}+ K_{2})\dot{h}_{i,j} + K_{1}K_{2}h_{i,j} \ge 0,
\end{equation}
where the time derivatives of $h_{i,j}$ are found by differentiating over the dynamics:
\begin{equation}
\label{eq: cbf filled in}
\begin{aligned}
    \dot{h}_{i,j} & = 2(\textbf{p}_{i}(0) - \textbf{p}_{j}(0)) \cdot (\textbf{v}_{i}(0) - \textbf{v}_{j}(0)) \\
    \ddot{h}_{i,j} & = 2 \lvert \lvert \textbf{v}_{i}(0) - \textbf{v}_{j}(0) \rvert \rvert^{2} \\
    & + 2(\textbf{p}_{i}(0) - \textbf{p}_{j}(0)) \cdot (\textbf{a}_{i}(0) - \textbf{a}_{j}(0)). \\ 
\end{aligned}
\end{equation}
This constraint should be repeated for all combinations of agent pairs. Note that the constraints are indeed linearly dependent on the acceleration input. The position and velocity variables of Eq.~(\ref{eq: cbf filled in}) are already known from the initial condition of OCP~(\ref{eq: ruben ADMM general form OCPX, filled in}). 

We use this HOCBF to formulate a convex quadratically constrained quadratic program (QCQP): 
\begin{equation}
\label{eq: CBF QP general form, filled in}
\begin{aligned}
    \min_{\forall \textbf{u}_i} \quad & \sum_{i=1}^{N} \lvert\lvert \textbf{u}_{i}(0) - \textbf{u}_{i}^{*}(0) \rvert \rvert^{2} \\
    \textrm{s.t.} \quad & \mathcal{H}_{i,j}(\textbf{u}_{i}(0), \textbf{u}_{j}(0)) \ge 0 \;\;,\quad  \forall j\neq i \\
    & \lvert \lvert \textbf{u}_{i}(0) \rvert \rvert^{2} \le a^2_{peak} \\
    & \forall i \in \{1,...,N\},
\end{aligned}
\end{equation}
where $\textbf{u}^{*}_{i}(0)$ denotes the control input at $t=0$ obtained by ADMM. Constraints~(\ref{eq: centralised ocp max acceleration}) are also added to ensure that the acceleration is still within the allowed range. To avoid infeasibilities, it can also be implemented as a soft constraint. However, the XPlanar has different power limits for continuous operation and short-term peak usage. Hence, we can increase the maximum allowed acceleration in QCQP~(\ref{eq: CBF QP general form, filled in}) ($a_{peak}$) w.r.t. its value in OCP~(\ref{eq: ruben ADMM general form OCPX, filled in}) ($a_{max}$). It is then possible to temporarily apply higher accelerations to avoid a collision, and QCQP~(\ref{eq: CBF QP general form, filled in}) will only do so if the original control input is not sufficient.

We compute QCQP~(\ref{eq: CBF QP general form, filled in}) after the ADMM iterations to obtain a hybrid method. Although decentralised CBFs can also be safe under certain conditions, centralised CBFs are more efficient and less conservative~\cite{borrmann2015cbfsafeswarm}. Since the XPlanar is controlled by a central computer, we can solve QCQP~(\ref{eq: CBF QP general form, filled in}) centrally. This choice is further motivated by the more tractable structure of this optimisation problem compared to OCP~(\ref{eq: Centralised ocp}).

\subsection{Pseudo-code}

We provide a pseudo-code of the hybrid method in Algorithm~\ref{alg: decentralised ADMM+CBF}. This algorithm takes every agent's initial state $\textbf{x}_{init,i}$ and target state $\textbf{x}_{f,i}$ as inputs. On lines 2-12, we perform $m$ ADMM iterations by repetitively solving Eq.~(\ref{eq: ruben ADMM general form OCPX, filled in})-(\ref{eq: ruben ADMM general form lambda update, filled in}) and communicating the intermediate results. Communication between agents is achieved by reading and writing these results in a shared memory since the system is controlled by a central computer. On line 14, we apply a safety filter to the inputs on the current time step by solving QCQP~(\ref{eq: CBF QP general form, filled in}). Finally, on lines 15-19, we simulate the system with the safe inputs using the agents' discrete-time dynamics and log the results. This procedure is repeated during every time step until all agents are sufficiently close to their target state.
The output of Algorithm~\ref{alg: decentralised ADMM+CBF} is the set of trajectories for all agents. The parameters that need to be tuned are the number of ADMM iterations per time step $m$, the penalty parameter $\mu$, the HOCBF coefficients $K_1$ and $K_2$, and the weights in the objective function $\textbf{Q}_{i}$ and $\textbf{R}_{i}$.

\begin{remark}
    Note that each ``for'' loop (lines 3 and 7) can be executed in parallel as there is no dependency on other agents within that for loop. However, a synchronisation step is needed after the first ``for'' loop (lines 3-6) because the results of all the agents are simultaneously needed for the second ``for'' loop (lines 7-11).
\end{remark}

\begin{remark}
    This algorithm can be hot-started by performing $m_{pre}$ ADMM pre-iterations before the first time step. In this way, the algorithm can start with a proper initial guess for $\textbf{z}_i(t)$, $\textbf{z}_{i,j}(t)$, $\boldsymbol{\lambda}_{i}(t)$ and $\boldsymbol{\lambda}_{i,j}(t)$. In practice, the pre-iterations could be performed while agents are standing still, e.g. when being loaded.
\end{remark}

\begin{algorithm}
    \caption{Hybrid method: ADMM-HOCBF}
    \label{alg: decentralised ADMM+CBF}
    \textrm{\textbf{Inputs}}: {$\textbf{x}_{init,i}$, $\textbf{x}_{f,i}$} \\
    \textrm{\textbf{Parameters}}: {$m, \mu, K_1, K_2, \textbf{Q}_{i}, \textbf{R}_{i}$} \\ 
    \textrm{\textbf{Outputs}}: {$\textbf{x}_{i}(t)$, $\textbf{u}_{i}(t)$}
    \begin{algorithmic}[1]
        \WHILE {$\lvert \lvert \textbf{x}_{c,i} - \textbf{x}_{f,i} \rvert \rvert^{2} \ge 10^{-3}$, $\: \forall i$}
            \FORALL{ADMM iterations $m$}
            \FORALL{agents $i \in \{1,...,N\}$}
                \STATE $\textbf{x}_{i}(t), \textbf{u}_{i}(t)$ according to OCP~(\ref{eq: ruben ADMM general form OCPX, filled in})
                \STATE Update $\textbf{p}_{i}(t)$ in shared memory 
            \ENDFOR
            \FORALL{agents $i \in \{1,...,N\}$}
                \STATE $\textbf{z}_{i}(t), \textbf{z}_{i,j}(t)$ according to NLP~(\ref{eq: ruben ADMM general form OCPZ, filled in})
                \STATE $\boldsymbol{\lambda}_{i}(t), \boldsymbol{\lambda}_{i,j}(t)$ according to Eq.~(\ref{eq: ruben ADMM general form lambda update, filled in})
                \STATE Update $\textbf{z}_{i}(t)$, $\textbf{z}_{i,j}(t)$, $\boldsymbol{\lambda}_{i}(t)$ and $\boldsymbol{\lambda}_{i,j}(t)$ in shared memory
            \ENDFOR
        \ENDFOR
        \STATE \COMMENT{Render input safe}
        \STATE $\textbf{u}_{1}(0)$, ..., $\textbf{u}_{N}(0)$ according to QCQP~(\ref{eq: CBF QP general form, filled in}) 
        \FORALL{agents $i \in \{1,...,N\}$}
            \STATE $\textbf{u}_{c,i} \gets \textbf{u}_{i}(0)$ 
            \STATE $\textbf{x}_{c,i} \gets \textbf{A}_{d}\textbf{x}_{c,i} + \textbf{B}_{d}\textbf{u}_{c,i}$
            \STATE $\textbf{x}_{0,i} \gets \textbf{x}_{c,i}$ 
            \STATE Log $\textbf{u}_{c,i}$ and $\textbf{x}_{c,i}$
        \ENDFOR
            \vspace{0.5em}
        \ENDWHILE
    \end{algorithmic}
\end{algorithm}

\subsection{Rectangular arena constraints}
\label{sec:rectangular arena constraints}

When operating on a transport system, agents must remain within the boundaries of this system. We start by considering a rectangular arena and formulate the following constraints on the agents' position:
\begin{equation}
\label{eq: box constraints general}
    \begin{aligned}
        & p_{x,min} + (\frac{w}{2} + \epsilon) \le p_{x,i}(t) \le p_{x,max} - (\frac{w}{2} + \epsilon) \\
        & p_{y,min} + (\frac{w}{2} + \epsilon) \le p_{y,i}(t) \le p_{y,max} - (\frac{w}{2} + \epsilon) \\
        & \forall i \in \{1,...,N\}.
    \end{aligned}
\end{equation}

\noindent
Here, $w$ denotes the width of an agent. Since the movers do not rotate during their trajectory, we model them as squares for this constraint, which is less conservative than the circular model for the collision avoidance constraints. These constraints must also be added to QCQP~(\ref{eq: CBF QP general form, filled in}). Since the positions are no longer decision variables here, we express Eq.~(\ref{eq: box constraints general}) as a function of the input:
\begin{equation}
\label{eq: box constraints general for CBF}
    \begin{aligned}
        & p_{x,min} + (\frac{w}{2} + \epsilon) \le p_{x,i}(0) + v_{x,i}(0) \Delta t + 0.5u_{x,i}(0) \Delta t^{2} \\
        & \le p_{x,max} - (\frac{w}{2} + \epsilon) \\
        & p_{y,min} + (\frac{w}{2} + \epsilon) \le p_{y,i}(0) + v_{y,i}(0) \Delta t + 0.5u_{y,i}(0) \Delta t^{2} \\
        & \le p_{y,max} - (\frac{w}{2} + \epsilon) \quad \quad \forall i \in \{1,...,N\}.
    \end{aligned}
\end{equation}

\noindent
We assume a constant acceleration law during each time step, making it compatible with the multiple shooting method, which will be used to solve Problems~(\ref{eq: ruben ADMM general form OCPX, filled in})-(\ref{eq: ruben ADMM general form OCPZ, filled in}) in Section~\ref{sec:experiments}.

\subsection{Corridor strategy for nonconvex arenas}
\label{sec: corridor stratgey}

\begin{figure}
    \centering
    \includegraphics[width=0.5\linewidth]{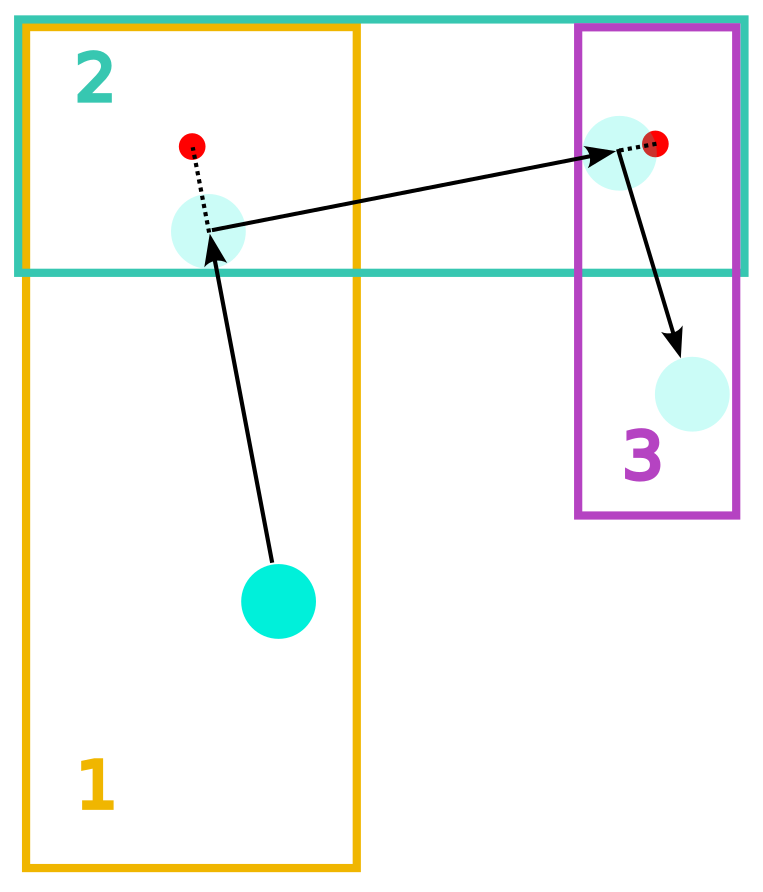}
    \caption{To deal with nonconvex arenas, agents track a sequence of sub-targets before arriving at their main target. The sub-targets are located in the intersection of corridors, where agents can switch from one corridor to the next. The final sub-target is the destination of the agent.}
    \label{fig:xplanar_corridors_flow}
\end{figure}

\noindent
The XPlanar arena consists of square tiles that can be configured in more general (nonconvex) shapes, as shown in Fig.~\ref{fig: xplanar experimental setup}. To deal with this, we use a heuristic corridor strategy, depicted in Fig.~\ref{fig:xplanar_corridors_flow}. First, the arena is partitioned into rectangular overlapping corridors. We then define a sub-target in each overlapping zone. If an agent's target is not in its current corridor, it must track a sequence of sub-targets. Every time the agent reaches an overlap zone, it immediately switches to the next sub-target. This procedure continues until the agent reaches the corridor with the final target. If the arena has a complex structure with many overlap zones, the sequence of sub-targets may be computed using a grid-based planner like A*.

This strategy ensures that every agent is always located in one corridor at a time, allowing the use of the rectangular arena constraints from Section~\ref{sec:rectangular arena constraints}. The structure of the constraints remains the same; only the dimensions are adaptively changed throughout the trajectory.

\begin{remark}
    In the current preliminary implementation, the sub-targets are placed in the middle of the overlap zone. In future research, we will adapt the position of the sub-targets for each agent according to their current scenario to avoid excessive clustering and make more efficient use of the overlap zone.
\end{remark}

%% file: content/section4.tex
\section{Results and experimental validation} \label{sec:experiments}

In this section, we present the results of our work, for which we provide an accompanying YouTube video\footnote{\label{YT video} https://youtu.be/5H8QRz0ndy4}. We first perform extensive Python simulations to compare ADMM-HOCBF to the classical centralised MPC approach based on scalability, performance, and safety. We also verify the deadlock resistance of ADMM-HOCBF. Then, we employ a fast C++ implementation to achieve real-time performance and conduct three different hardware experiments by deploying the calculated trajectories on a Beckhoff XPlanar system. 

\subsection{Simulation results}
In simulations, we use the open-source toolkits CasADi \cite{andersson2019casadi} and Rockit \cite{gillis2020rockit} to configure and transcribe the three optimisation problems (OCP~(\ref{eq: ruben ADMM general form OCPX, filled in}), NLP~(\ref{eq: ruben ADMM general form OCPZ, filled in}), and QCQP~(\ref{eq: CBF QP general form, filled in})) in Python. For transcription, we use the multiple shooting method with sampling time $\Delta t = 100$ ms. We use IPOPT \cite{wachter2006ipopt} to solve all optimisation problems to ensure a fair comparison between the methods. All tests are performed on a Linux computer with a 7th Gen Intel\textregistered \space Core\texttrademark \space i7-7700K CPU @ 4.20GHz. Finally, the ADMM-HOCBF scheme is still implemented sequentially in Python; the ability to parallelise Algorithm~\ref{alg: decentralised ADMM+CBF} has not yet been exploited.

\subsubsection{Scalability analysis}
 
We test 17 scenarios varying from 2 to 30 agents. For each scenario, we perform 20 simulations using random start and target positions for each agent. Each case is solved using three different methods: (I) centralised MPC, (II) ADMM-HOCBF with $m=1$ and $\mu = 1$, and (III) ADMM-HOCBF with $m=20$ and $\mu = 40$. The other parameters are fixed: $T_f = 1$ s, $\Delta t = 100$ ms, $m_{pre} = 50$, $K_1 = 8 \: \text{s}^{-1}$, $K_2 = 7 \: \text{s}^{-1}$, $\textbf{Q}_{i} = diag([1 \: \text{m}^{-2}, 1 \: \text{m}^{-2}, 0 \: (\text{m/s})^{-2}, 0 \: (\text{m/s})^{-2}])$, $\textbf{R}_{i} = 10^{-4}\times I_{2 \times 2} \: (\text{m/s$^2$})^{-2}$ , $v_{max} = 1 \: \text{m/s}$, $a_{max} = 5 \: \text{m/s}^2$, $a_{peak} = 8 \: \text{m/s}^2$. 

\begin{remark}
    Since we operate at rather high velocities, the safety margin for centralised MPC has to be set at $\epsilon = 30$ mm to avoid safety violations between the grid points after discretisation. For ADMM-HOCBF, we can strongly reduce the safety margin to $\epsilon = 5$ mm since the HOCBF tends to slow down agents as they approach each other.
\end{remark}

\begin{figure}
    \centering
    \includegraphics[width=1\linewidth]{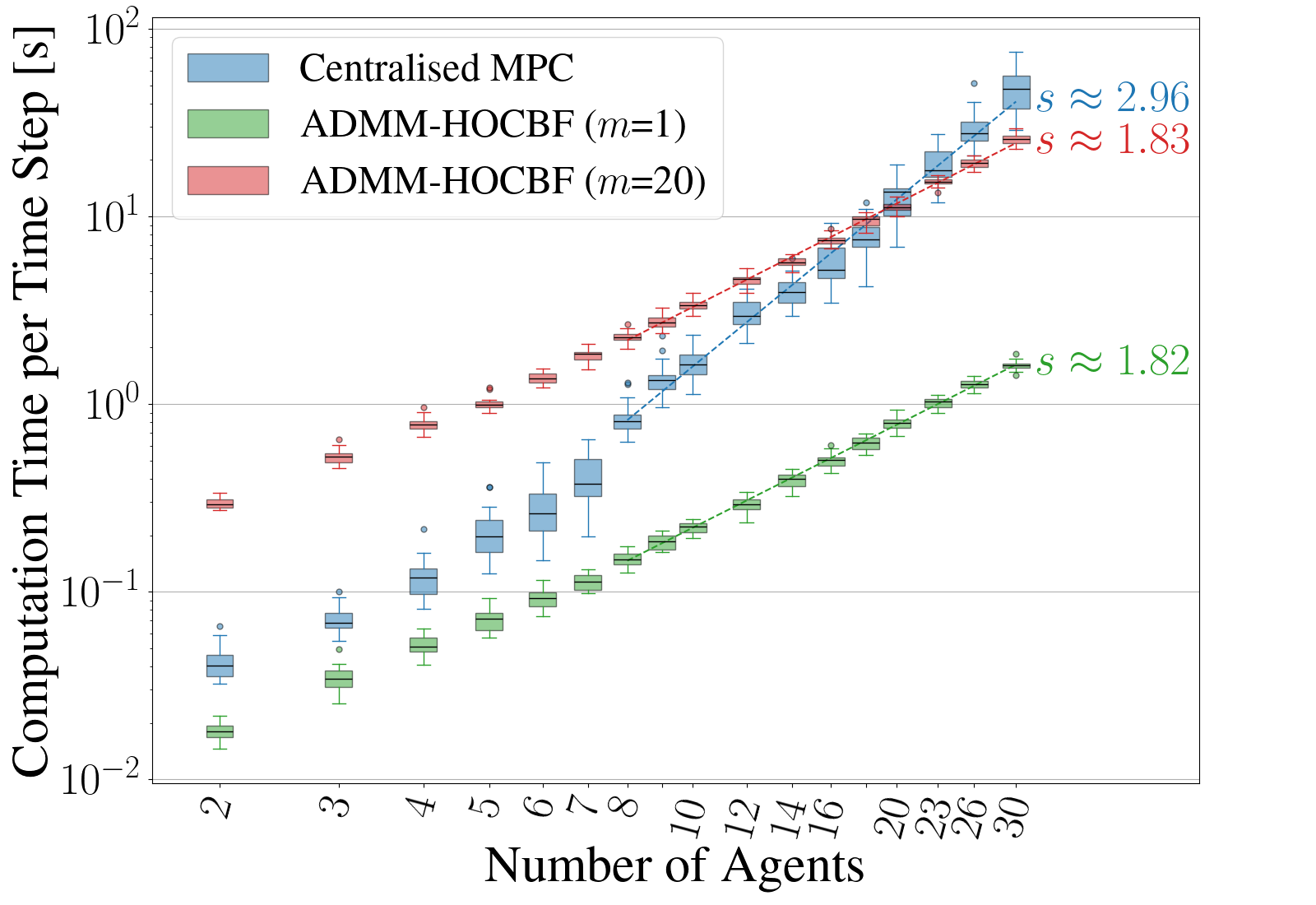}
    \caption{Scalability plot of ADMM-HOCBF ($m=1$ and $m=20$) and centralised MPC.}
    \label{fig:scalability}
\end{figure}

We log the computation time per time step for each instance and illustrate the results in Fig.~\ref{fig:scalability} using three distinct colours. We notice a sharp difference in scalability between centralised MPC ($\sim O(N^{2.96})$) and ADMM-HOCBF ($\sim O(N^{1.83})$). For the latter, the scalability is approximately independent of the number of ADMM iterations $m$. Another advantage is a lower uncertainty in the computation time per time step. This is due to the fact that ADMM-HOCBF does not have to compute to full convergence during each time step unlike centralised MPC, and it can be done without losing safety guarantees.

\subsubsection{Performance comparison and safety}
\label{sec:performance comparison}
The same three methods are compared for one case with five agents in terms of the transit time, average computation time per time step, and number of collisions. We also investigate the importance of the HOCBFs by calculating percentage-wise how many times, and by how much, the safety filter has to adjust the input along the trajectory. The results are shown in Tab.~\ref{tab:comp_admmcbf_centrmpc}. 

\begin{table}[h]
    \centering
    \begin{tabular}{|c|c|c|c|}
    \hline
         & \multicolumn{2}{c|}{ADMM-HOCBF} & Centr. MPC \\
    \cline{2-3}
         & $m=1$ & $m=20$ &  \\
    \hline
       transit time  & $4.6$ s & $3.0$ s & $2.0$ s \\
    \hline
        avg. computation time & $115$ ms & $1244$ ms & $169$ ms \\
    \hline
        number of collisions & $0$ & $0$ & $0$ \\
    \hline
        safety filter activity & $65.22\%$ & $36.67\%$ & $-$ \\
    \hline
    avg. input correction & $6.75 \; \text{m/s}^{2}$ & $1.80 \: \text{m/s}^{2}$ & $-$ \\
    \hline
    \end{tabular}
    \caption{Performance and safety comparison between sequential ADMM-HOCBF and centralised MPC. Tests are performed in Python for five agents.}
    \label{tab:comp_admmcbf_centrmpc}
\end{table}

ADMM-HOCBF with $m=1$ produces the least efficient trajectories, but exhibits the lowest computational cost. Increasing the number of ADMM iterations reduces the transit time, but it also increases the computation time from $115$ ms to $1244$ ms. This might seem detrimental, as the computation time is now larger than that of centralised MPC, but from around 20 agents, the computation time of the latter surpasses ADMM-HOCBF (with $m=20$), as shown in Fig.~\ref{fig:scalability}. Hence, the computational cost starts at a higher value but increases more gently.

The number of collisions is zero for every method, confirming that the trajectories are safe. We observe that ADMM-HOCBF with $m=1$ relies more heavily on the safety filter than ADMM-HOCBF with $m=20$. This results in more reactive behaviour, which explains its larger transit time. Because a rising number of ADMM iterations per time-step results in trajectories with a lower transit time but higher computation time, the efficiency of the trajectories can be directly linked to the available hardware on which the algorithm is executed. 

\subsubsection{Deadlocks}
Since ADMM-HOCBF produces reactive behaviour at a lower number of ADMM iterations, we verify its performance in a case which is susceptible to deadlocks. The result is shown in the video\footref{YT video}. Although the agents' reactive behaviour causes a deadlock, they continue performing ADMM iterations, improving the approximate solution of the OCP, which eventually allows them to escape the deadlock. In future research, we will investigate this deadlock resistance quantitively.

\subsection{Experimental results}
\label{sec: exp res}

To verify whether the trajectories produced by Algorithm~\ref{alg: decentralised ADMM+CBF} are suitable for real industrial applications, they are tested on an XPlanar setup, shown in Fig.~\ref{fig: xplanar experimental setup}. The improved scalability of ADMM-HOCBF motivates the development of a highly efficient multithreaded C++ code with more tailored solvers. We use FATROP \cite{vanroye2023fatrop} for problems~(\ref{eq: ruben ADMM general form OCPX, filled in})-(\ref{eq: ruben ADMM general form OCPZ, filled in}), and qpOASES \cite{ferreau2014qpoases} for QCQP~(\ref{eq: CBF QP general form, filled in}). We conduct 3 experiments with 5 movers, which are also shown in the video\footref{YT video}. The first two experiments are performed within the largest rectangle of the arena, while the last uses the entire arena.

\subsubsection{Performance analysis}

\begin{figure} [h]
    \centering

    \begin{subfigure}[b]{0.35\textwidth}
        \includegraphics[width=\linewidth, trim={2.5cm 10cm 5cm 5cm},clip]{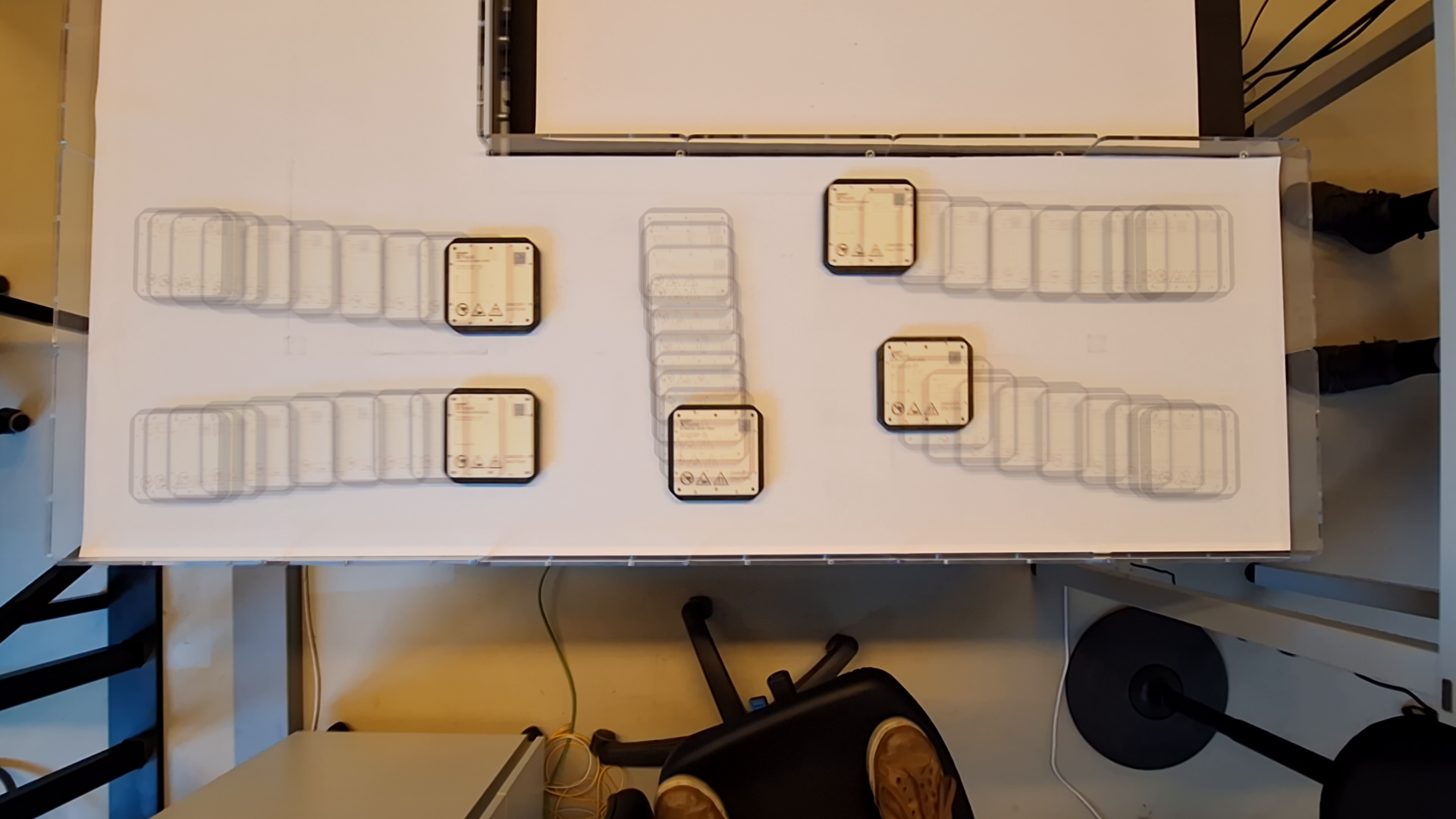}
        \caption{$t=0-1\:s$}
    \end{subfigure}
    \vspace{0.2cm}
    
    \begin{subfigure}[b]{0.35\textwidth}
        \includegraphics[width=\linewidth, trim={2.5cm 10cm 5cm 5cm},clip]{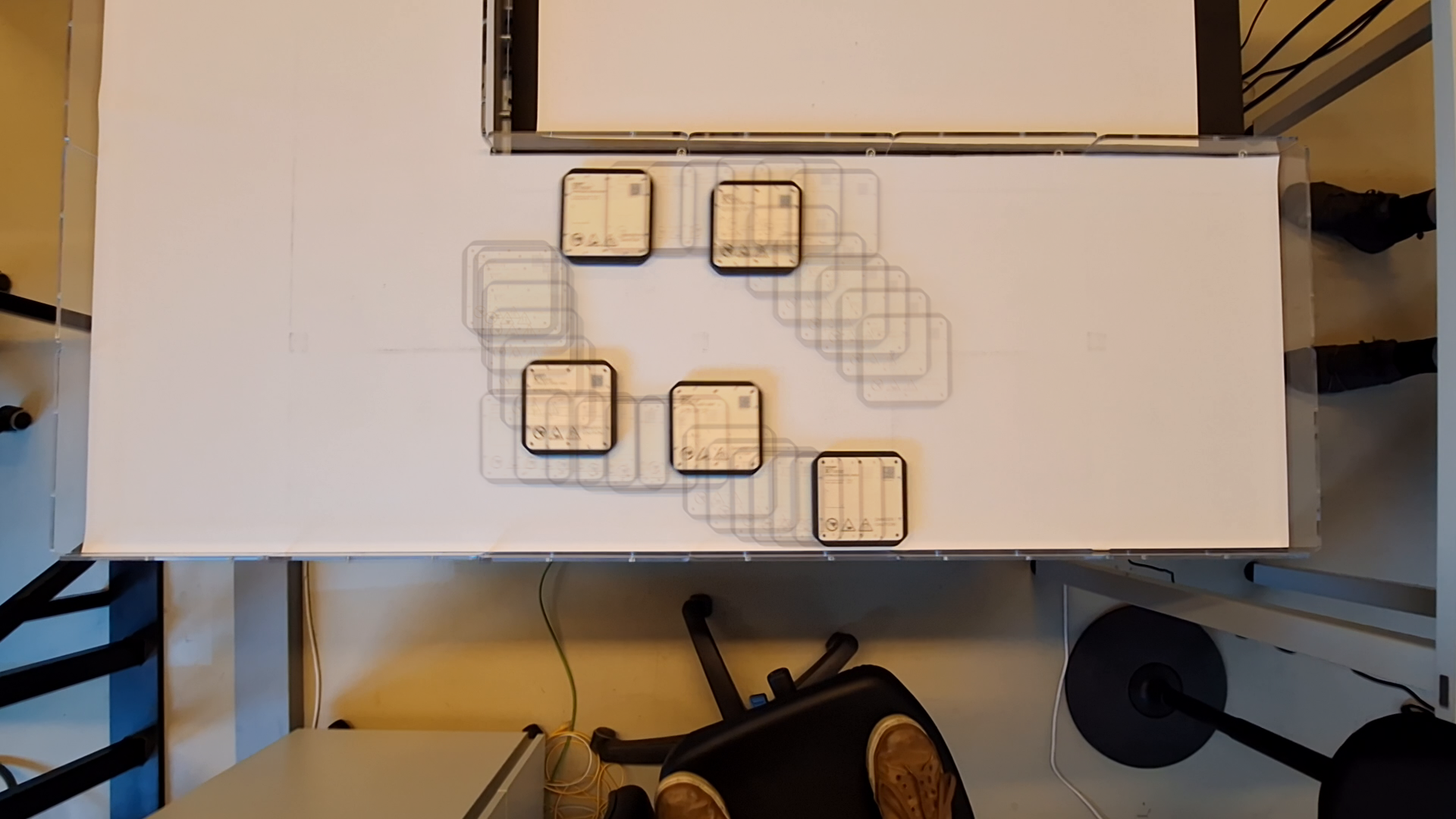}
        \caption{$t=1-1.6\:s$}
    \end{subfigure}
    \vspace{0.2cm}
   
    \begin{subfigure}[b]{0.35\textwidth}
        \includegraphics[width=\linewidth, trim={2.5cm 10cm 5cm 5cm},clip]{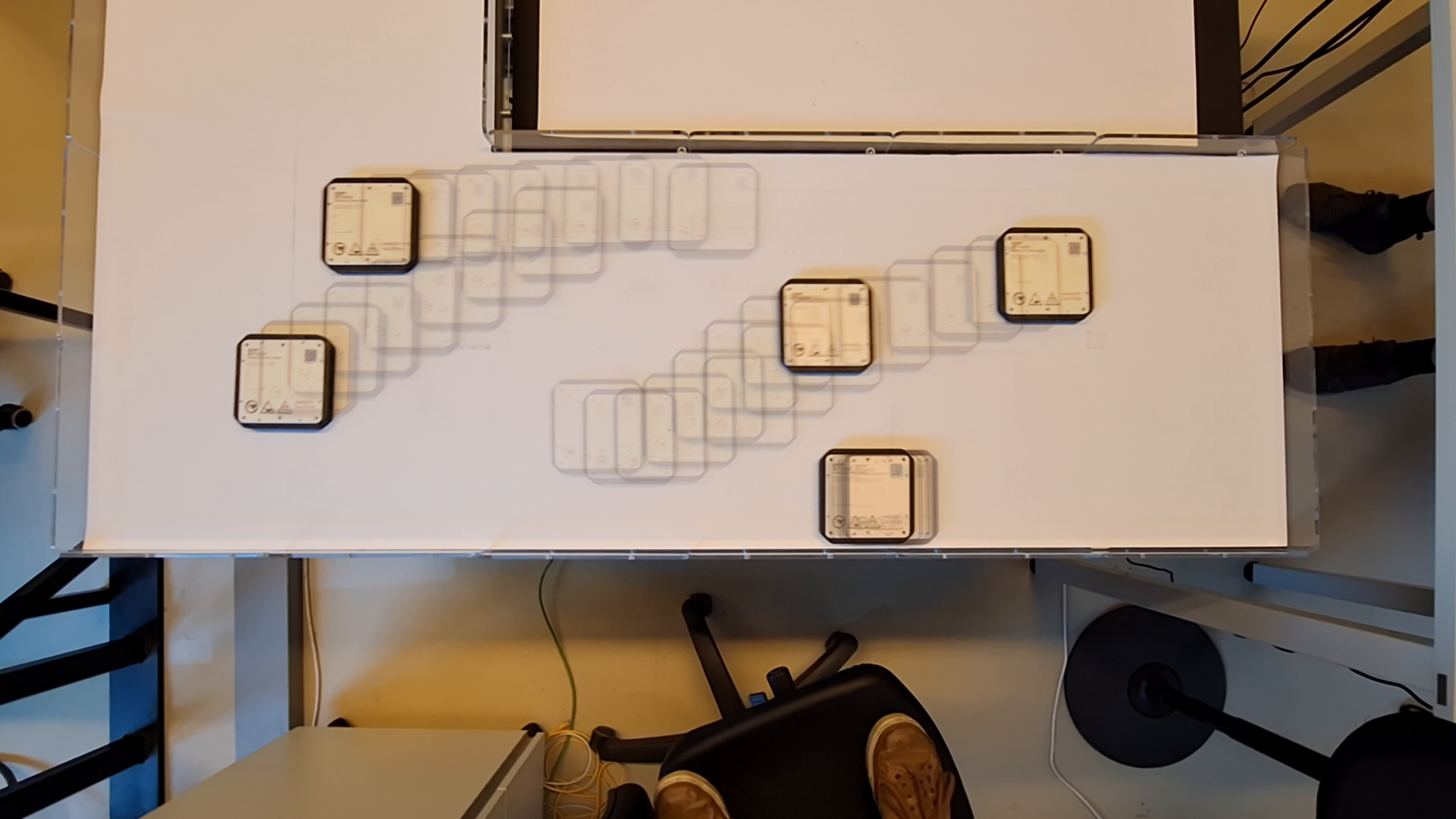}
        \caption{$t=1.6-2.4\:s$}
    \end{subfigure}
    \vspace{0.2cm}
   
    \begin{subfigure}[b]{0.35\textwidth}
        \includegraphics[width=\linewidth, trim={2.5cm 10cm 5cm 5cm},clip]{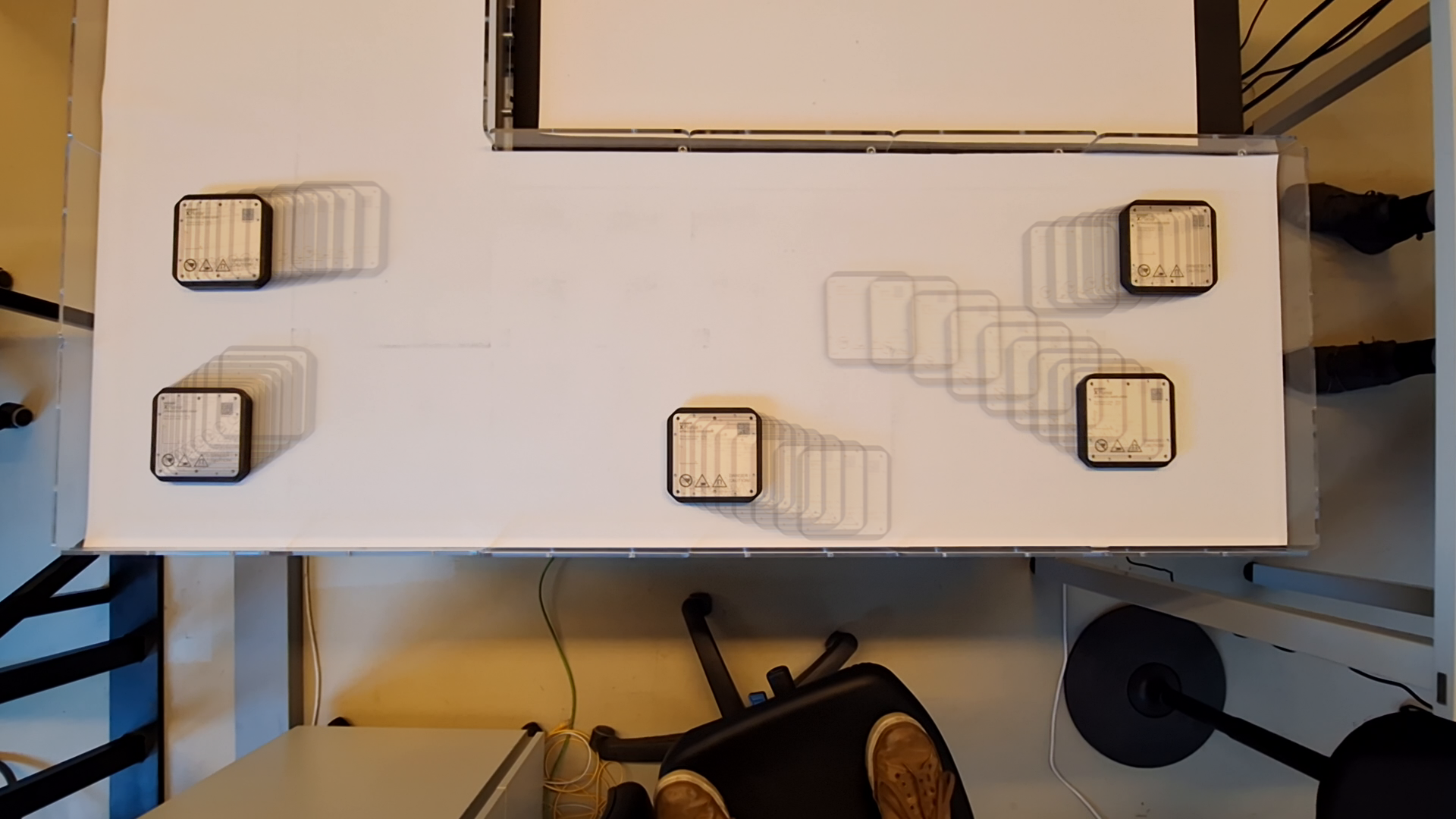}
        \caption{$t=2.4-3.6\:s$}
    \end{subfigure}

    \caption{Trajectories of 5 agents on the XPlanar system with ADMM-HOCBF (m=20).}
    \label{fig: hardware trajectories}
\end{figure}

In the first experiment, we compute trajectories for 5 XPlanar movers using ADMM-HOCBF with $m=1$ and $m=20$. To restrict these movers to the main rectangle of the setup, the (tight) arena constraints from Section~\ref{sec:rectangular arena constraints} must be included. Fig.~\ref{fig: hardware trajectories} shows the resulting trajectories for $m=20$. The experimental results are consistent with the simulations, yielding the same trade-off between efficiency and computation time. However, the computation time per time step is significantly improved and ADMM-HOCBF with $m=1$ exhibits real-time performance by a large margin, as shown in Table~\ref{tab:comp_admmcbf_centrmpc_experimental}.

\begin{table}[h]
    \centering
    \begin{tabular}{|c|c|c|}
    \hline
         & \multicolumn{2}{c|}{ADMM-HOCBF}\\
    \cline{2-3}
         & $m=1$ & $m=20$ \\
    \hline
       transit time  & $4.5$ s & $3.6$ s \\
    \hline
        avg. computation time & $20$ ms & $204$ ms \\
    \hline
    \end{tabular}
    \caption{Performance results of the parallel C++ implementation of ADMM-HOCBF.}
    \label{tab:comp_admmcbf_centrmpc_experimental}
\end{table}

\subsubsection{Continuous operation}
In the second experiment, every agent starts by tracking its own target. Whenever it reaches this target, it is instantly assigned a new random one. This process continues for 25 s, at which point the command is given to all agents to return to their starting positions, immediately overwriting the targets they were originally tracking. This experiment shows that the motion planner is capable of working with changing targets, making it flexible, efficient, and reliable for continuous operation.

\subsubsection{General arena shape}
In the final experiment, we allow targets to be spread over the full non-convex arena. This is achieved by employing the corridor strategy, highlighted in Section~\ref{sec: corridor stratgey}. This experiment shows that the algorithm can be implemented in complex realistic environments.

%% file: content/section5.tex
\section{Conclusions and future work} 
\label{sec:conclusion}

This paper presents an MPC-based motion planner that utilises a combination of decentralised ADMM and centralised HOCBFs. ADMM-HOCBF is shown to scale better than centralised MPC, without losing safety guarantees. Experimental validation of the algorithm is conducted on the Beckhoff XPlanar system, where ADMM-HOCBF is shown to be promising for real-time applications. It is also shown that the algorithm can be extended to deal with nonconvex arenas by partitioning the arena into corridors.

Future work consists of three parts. For practical use, integrating a more efficient corridor strategy and a framework for online computations is essential. To further improve performance, we can explore ADMM acceleration techniques \cite{goldstein2014fast} and dynamic parameter tuning. To further improve scalability, possibilities include introducing more efficient, time-varying neighbourhoods, which is non-trivial for ADMM-based algorithms \cite{tian2022timevarying}, and incorporating learning techniques into the non-safety-critical parts of the algorithm.

%% file: content/section6.tex
\section{Acknowledgements} 
\label{sec:acknowledgements}

The authors would like to thank Joris Gillis for his support with CasADi in the C++ implementation.